# Optimizing Semiconductor Devices by Self-organizing Particle Swarm

Xiao-Feng Xie
Institute of Microelectronics,
Tsinghua University,
Beijing 100084, P. R. China
Email:xiexf@ieee.org

Wen-Jun Zhang
Institute of Microelectronics,
Tsinghua University,
Beijing 100084, P. R. China
Email: zwj@tsinghua.edu.cn

De-Chun Bi
Department of Environmental Engineering,
Liaoning University of Petroleum
& Chemical Technology
Fushun, Liaoning, 113008, P. R. China

*Abstract*- **A self-organizing particle swarm is presented. It works in dissipative state by employing the small inertia weight, according to experimental analysis on a simplified model, which with fast convergence. Then by recognizing and replacing inactive particles according to the process deviation information of device parameters, the fluctuation is introduced so as to driving the irreversible evolution process with better fitness. The testing on benchmark functions and an application example for device optimization with designed fitness function indicates it improves the performance effectively.**

## I. INTRODUCTION

From the deep sub-micrometer to nanoscale devices, the technology CAD (TCAD) tools provide a better insight than measurement techniques and have become indispensable in the new device creation [4]. Technology development, however, requires substantially more than fundamental simulation capability: tools and methods to assist in exploration of design trade-offs and to optimize a design, are becoming increasingly important [6].

As the first step in technology synthesis [6, 8], device optimization problem is to finding some feasible device designables in the design space (S) in order to satisfying the requirements on the device electrical performance. Each device designable $\vec{x} = (x_1,...,x_d,...,x_D) \in S$ is a possible combination of device parameters (xd) that specify the device topography and channel impurity concentrations associated with a semiconductor device. Typical device designables are physical gate length, oxide thickness, etc. The D-dimensional design space (S) is a Cartesian product of domains of device parameters. Device performance is evaluated from the electrical responses of the device. Examples are on and off currents, threshold voltage, output resistance, etc. for MOS devices.

For short channel MOS devices, the response surfaces of device performance are often highly nonlinearly. For scaled devices, such surfaces can be even highly rugged due to the disturbance from the mesh adjustment of device simulations, such as PISCES [22]. Moreover, the highly sophisticated novel semiconductor devices involve many design parameters, which increase the dimension of design space. For handling with such complex cases, the global search techniques, such as genetic algorithm (GA) [10], particle swarm optimization (PSO) [5, 7], etc., had been employed in a modeling and optimization system [21] for finding the feasible devices.

The number of evaluation times is crucially for device optimization since each device simulation is time-consumptively. As a novel stochastic algorithm, PSO [5, 7] is inspired by social behavior of swarms. Studies [1] showed that although PSO discovered reasonable quality solutions much faster than other evolutionary algorithms (EAs), however, it did not possess the ability to improve upon the quality of the solutions as evolution goes on.

Structures of increasing complexity in self-organizing dissipative systems based on energy exchanges with the environment were developed into a general concept of dissipative structures (DS) by Prigogine [10, 12], which allows adaptation to the prevailing environment with extraordinarily flexible and robust. The self-organization of DS was implemented in a dissipative PSO (DPSO) model [19] with some good results. However, the DPSO model also introduces additional control parameters to be adjusted by hand, which cannot be very easily determined.

In this paper, the particle swarm is worked in dissipative state with the control parameters according to the experimental convergence analysis results from a simplified model. Instead of introducing additional parameters, the process deviations of device parameters are employed for recognizing inactive particles. Then the negative entropy is introduced to stimulate the model of PSO operating as a DS, which is realized by replacing inactive particles with fresh particles. The variant of particle swarm, termed self-organizing PSO (SOPSO), removes the additional parameters. The performance of the self-organizing PSO is studied on three benchmark functions and an application for a Focused-Ion-Beamed MOSFET (FIBMOS) [13] with designed fitness function, by comparing with some existing algorithms [9, 14, 19].

## II. PARTICLE SWARM OPTIMIZATION

In particle swarm, The location of the *i*th ($1 \le i \le N$, $i \in \mathbb{Z}$) particle is a potential solution in *S* represented as $\vec{x}_i = (x_{i1},...,x_{id},...,x_{iD})$. The best previous position (with the best fitness value) of the *i*th particle is recorded and represented as $\vec{p}_i = (p_{i1},..., p_{id}, ..., p_{iD})$, which is also called *pbest*. The index of the best *pbest* among all the particles is represented by the symbol *g*. The $\vec{p}_g$ (or $\vec{g}$) is also called *gbest*. The velocity for the *i*th particle is $\vec{v}_i = (v_{i1},..., v_{id}, ..., v_{iD})$. At each time step, the *i*th particle is manipulated according to the following equations [14]:

$$v_{id} = w \cdot v_{id} + c_1 \cdot U_\mathbb{R}() \cdot (p_{id} - x_{id}) + c_2 \cdot U_\mathbb{R}() \cdot (g_d - x_{id}) \quad (1a)$$
$$x_{id} = x_{id} + v_{id} \quad (1b)$$

where $w$ is inertia weight, $c_1$ and $c_2$ are acceleration constants, $U_\mathbb{R}()$ are random values between 0 and 1.

## III. EXPERIMENTAL CONVERGENCE ANALYSIS

The convergence analysis can be done on a single dimension without loss of generality since there is no interaction between the different dimensions in equations (1), so that the subscript $d$ is dropped. And the analysis is simplified even more for a single particle, then the subscript $i$ is dropped, i.e.,

$$v = w \cdot v + c_1 \cdot U_\mathbb{R}() \cdot (p - x) + c_2 \cdot U_\mathbb{R}() \cdot (g - x) \quad (2a)$$
$$x = x + v \quad (2b)$$

Here just consider a special case $|p-g| \to 0$, which can be encountered by every particle. By transforming the coordinate origin to $g$, i.e. $p \approx g = 0$, equation (2a) can be simplified as:

$$v = w \cdot v - C \cdot x \quad (3)$$

where $C = (c_1 \cdot U_\mathbb{R}() + c_2 \cdot U_\mathbb{R}())$.

Several researchers have analyzed theoretically for the cases when $C$ is fixed [2, 16, 17]. However, it still has great difficulty to analyze the cases when $C$ is stochastically varied. Without loss of generality, here we suppose the $g$ is in a local minimum, i.e. is not changed in the following generations. Fig. 1 gives the average $\log(|x|)$ value of 1E6 experimental trails in 100 generations, where $\log()$ is the logarithmic operator, the initial $x^{(0)} = U_\mathbb{R}()$, the initial $v^{(0)} = U_\mathbb{R}()$, $c_1 = c_2 = 2$. Besides, a special setting that in constriction factor (CF) [2] is also demostrated in Fig.1 as a dash line, which with $w = 0.729$ and $c_1 = c_2 = 1.494$. Here it can be seen that the average $\log(|x|)$ varied linearly as generation $t$ is increasing. Besides, the CF version lies between $w = 0.6 \sim 0.8$ (about 0.65) in the normal version.

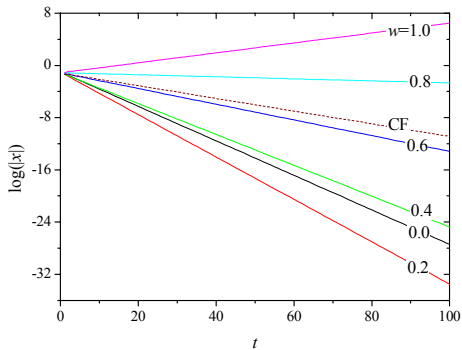

Figure 1. The average $\log(|x|)$ for different $w$ in 100 generations.

Fig. 2 gives the average $\log(|x|)$ for different $w$ in $T=100$ generations. The swarm is in chaotic state when $w > w_{th}$, and be dissipative state when $0 \le w \le w_{th}$, here $w_{th}$ is about equal to 0.85. Of course, for real swarm, such threshold value should be decreased slightly since normally $p \ne g$ at the initial time. The value of $w$ that provides balance between exploitation and exploration is lower than the $w_{th}$, which is about $0.5 \sim 0.75$ [11, 19], since the performance of swarm decreases when it be too dissipative or be too chaos. Moreover, it can be seen there has a higher tail when $w$ is closed to zero, which is agreed with the increasing fitness at the tail of the experimental results [19] for Rastrigrin and Griewank functions.

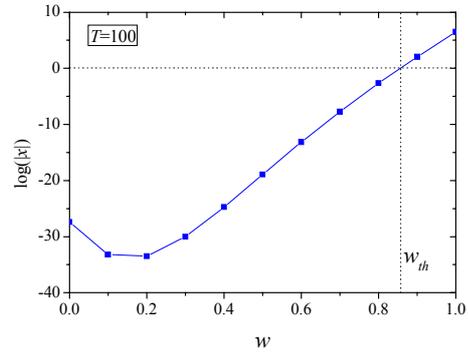

Figure 2. The average $\log(|x|)$ for different $w$ at $T=100$.

## IV. SELF-ORGANIZING PARTICLE SWARM

To converge fast with considerable performance, a self-organizing PSO (SOPSO) based on the principle of DS is presented, as shown in Fig. 3.

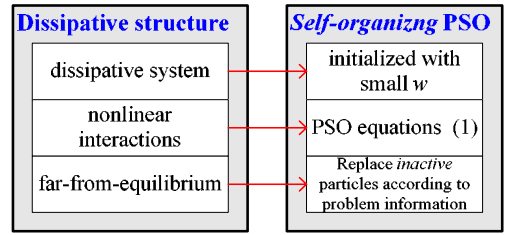

Figure 3. Principle of self-organizing PSO: performing as a DS.

Firstly, as a prerequisite of DS [10, 12], small $w$, which is much less than $w_{th}$, is adopted to induce the dissipative processes into the particle swarm, i.e. to reduce the average energy of swarm during evolution according to the nonlinear interactions among particles.

Secondly, the self-organization requires a system consisting of multiple elements in which nonlinear interactions between system elements are present [10]. In PSO, the equation (1) ensures nonlinear relations of positive and negative feedback between particles according to the cooperation and competition with the particle with best experience so far.

Thirdly, far-from-equilibrium conditions should be taken for granted [12]. As time goes on, some particles become *inactive* [20] when they are similar to the *gbest* and lost their velocities, which lost both of the global and local search capability in the next generations. Since the inactive particle has less contribution on evolution results, it suggests an adaptive mechanism to introduce negative entropy as following: recognize and replace it.

Taking the problem information into account can improve the algorithm performance [18]. For each device, there exists process deviation $\sigma_d$ for each parameter $x_d$. For a given point $\vec{o}$, its *similar set* $S_s$ is defined as all the points within the range of the deviations. Then for $\forall \vec{x} \in S_s$, which $|x_d - o_d| \leq \sigma_d$ come into existence in all dimensions, is noted as $\vec{x} \cong \vec{o}$, i.e. $\vec{x}$ is similar to $\vec{o}$. Then a particle can be identified as *inactive* if it is always flying in $S_s$ of $\vec{p}_g$ and without any improvement on its fitness in the following $T_c$ generations.

The pseudo code for recognizing inactive particle is shown in Fig. 4, which is executed after equations (1). The *SC* array is employed to store the times that are similar to the *gbest* in successively for each particle. $T_c$ is a predefined constant. It need not be too large since the swarm dynamics is dissipative. According to our experience, $T_c$=2 is good enough.

```
IF ( x⃗_i ≅ g⃗ && i!=g)      // similar to the gbest?
   THEN SC[i] ++;           // increase 1
ELSE SC[i] = 0;             // reset to 0
IF (SC[i]>T_c)              // achieve predefined count?
   THEN the ith particle is inactive;
```

Figure 4. Pseudo code for recognizing an *inactive* particle.

If the *i*th particle is recognized as inactive, then it is replaced by a *fresh* particle, which the location $\vec{x}_i$ and $\vec{v}_i$ is reinitialized at random, and *SC*[*i*] is reset to 0.

## V. FITNESS FUNCTION DESIGN

Typical requirements on device responses are classified in two classes of objectives:

$$\begin{cases} \text{minimize}: f_j(\vec{x}), & \text{for } j \in [1, m_0] \quad \text{(MIN type)} \\ g_k(\vec{x}) \in [c_{l,k}, c_{u,k}], & \text{for } k \in [m_0+1, m] \quad \text{(CON type)} \end{cases} \quad (4)$$

where $f$ and $g$ are response functions calculated by device simulator. The number of functions in MIN type often are normally 0 or 1 [8, 21]. The set of points, which satisfying all functions $g_k$ in CON type, is denoted as feasible space ($S_F$).

The total fitness function $F(\vec{x})$ is defined as:

$$F(\vec{x}) = <F_{OBJ}(\vec{x}), F_{CON}(\vec{x})> \quad (5)$$

where $F_{OBJ}(\vec{x}) = \sum_{j=1}^{m_0} w_j f_j(\vec{x})$ and $F_{CON}(\vec{x}) = \sum_{k=m_0+1}^{m} G_k(\vec{x})$ are the fitness functions for objectives and constraints, respectively, where $w_j$ are positive weight constants, which default values are equal to 1, and

$$G_k(\vec{x}) = \begin{cases} 0 & g_k(\vec{x}) \in [c_{l,k}, c_{u,k}] \\ (c_{l,k} - g_k(\vec{x}))/A_k & g_k(\vec{x}) < c_{l,k} \\ (g_k(\vec{x}) - c_{u,k})/B_k & g_k(\vec{x}) > c_{u,k} \end{cases} \quad (6)$$

where $A_k = \begin{cases} |c_{l,k}| & c_{l,k} \neq 0 \\ 1 & c_{l,k} = 0 \end{cases}$ and $B_k = \begin{cases} |c_{u,k}| & c_{u,k} \neq 0 \\ 1 & c_{u,k} = 0 \end{cases}$ are used for normalizing the large differences in the magnitude of the constraint values for device performance.

In order to avoid the difficulty to adjusting penalty coefficient in penalty function methods [9], the fitness evaluation is realized by directly comparison between any two points $\vec{x}_A$, $\vec{x}_B$, i.e. $F(\vec{x}_A) \leq F(\vec{x}_B)$ when

$$\begin{cases} F_{CON}(\vec{x}_A) < F_{CON}(\vec{x}_B) \text{ OR} \\ F_{OBJ}(\vec{x}_A) \leq F_{OBJ}(\vec{x}_B), F_{CON}(\vec{x}_A) = F_{CON}(\vec{x}_B) \end{cases} \quad (7)$$

If $F_{CON}$=0, then $\vec{x} \in S_F$. It is also following the Deb's criteria [3]: a) any $\vec{x} \in S_F$ is preferred to any $\vec{x} \notin S_F$; b) among two feasible solutions, the one having better $F_{OBJ}$ is preferred; c) among two infeasible solutions, the one having smaller $F_{CON}$ is preferred.

The device simulator may also be failed to calculate the responses for some designables when meets with inappropriate mesh settings. Then for such wrong cases, $F_{OBJ} = F_{CON} = +\infty$.

## VI. RESULTS AND DISCUSSION

### A. Benchmark functions

For comparison, three unconstrained benchmark functions that are commonly used in the evolutionary computation literature [14] are used. It can be done loss of generality, since the $F(\vec{x})$ of constrained function can be seemed as a uncostrained function. All functions have same minimum value, which are equal to zero.

The function $f_1$ is the Rosenbrock function:

$$f_1(\vec{x}) = \sum_{d=1}^{D-1} (100(x_{d+1} - x_d^2)^2 + (x_d - 1)^2) \quad (8a)$$

The function $f_2$ is the generalized Rastrigrin function:

$$f_2(\vec{x}) = \sum_{d=1}^{D} (x_d^2 - 10\cos(2\pi x_d) + 10) \quad (8b)$$

The function $f_3$ is the generalized Griewank function:

$$f_3(\vec{x}) = \frac{1}{4000} \sum_{d=1}^{D} x_d^2 - \prod_{d=1}^{D} \cos\left(\frac{x_d}{\sqrt{d}}\right) + 1 \quad (8c)$$

For $d$th dimension, $x_{max,d}$=100 for $f_1$, $x_{max,d}$=10 for $f_2$; $x_{max,d}$=600 for $f_3$. Acceleration constants $c_1=c_2=2$. The fitness value is set as function value. We had 500 trial runs for every instance.

Table 1 lists the initialization ranges. Table 2 gives the additional test conditions. FPSO give the results of fuzzy adaptive PSO [14]. DPSO give the results of PSO version in, which $c_v$=0, $c_f$=0.001 [19]. For SOPSO, $w$=0.4, and $\sigma_d$=0.01 for the $d$th dimension of all functions. In order to investigate whether the SOPSO scales well or not, different numbers of particles ($N$) are used for each function which different dimensions. The numbers of particles $N$ are 20, 40 and 80. The maximum generations $T$ is set as 1000, 1500 and 2000 generations corresponding to the dimensions 10, 20 and 30, respectively.

TABLE 1. INITIALIZATION RANGES

| Function | Asymmetric | Symmetric |
|---|---|---|
| $f_1$ | (15,30) | (-100, 100) |
| $f_2$ | (2.56,5.12) | (-10,10) |
| $f_3$ | (300,600) | (-600,600) |

TABLE 2. TEST CONDITIONS FOR DIFFERENT PSO VERSIONS

| PSO Type | FPSO [14] | DPSO [19] | SOPSO |
|---|---|---|---|
| Initialization | Asymmetric | Symmetric | Symmetric |
| $w$ | Fuzzy | 0.4 | 0.4 |

TABLE 3. THE MEAN FITNESS VALUES FOR THE ROSENBROCK FUNCTION

| $N$ | $D$ | $T$ | FPSO [14] | DPSO | SOPSO |
|---|---|---|---|---|---|
| 20 | 10 | 1000 | 66.01409 | 35.7352 | 21.7764 |
|  | 20 | 1500 | 108.2865 | 78.5368 | 47.1619 |
|  | 30 | 2000 | 183.8037 | 132.1512 | 72.1907 |
| 40 | 10 | 1000 | 48.76523 | 17.0553 | 13.4052 |
|  | 20 | 1500 | 63.88408 | 43.8963 | 29.2476 |
|  | 30 | 2000 | 175.0093 | 82.7209 | 52.2661 |
| 80 | 10 | 1000 | 15.81645 | 17.7516 | 11.9570 |
|  | 20 | 1500 | 45.99998 | 32.2961 | 26.5855 |
|  | 30 | 2000 | 124.4184 | 57.2802 | 46.9986 |

TABLE 4. THE MEAN FITNESS VALUES FOR THE RASTRIGRIN FUNCTION

| $N$ | $D$ | $T$ | FPSO [14] | DPSO [19] | SOPSO |
|---|---|---|---|---|---|
| 20 | 10 | 1000 | 4.955165 | 0.4707 | 1.0079 |
|  | 20 | 1500 | 23.27334 | 2.5729 | 6.8493 |
|  | 30 | 2000 | 48.47555 | 7.3258 | 17.7011 |
| 40 | 10 | 1000 | 3.283368 | 0.0762 | 0.1576 |
|  | 20 | 1500 | 15.04448 | 1.3088 | 3.2670 |
|  | 30 | 2000 | 35.20146 | 6.2107 | 9.5294 |
| 80 | 10 | 1000 | 2.328207 | 0.0080 | 0.0086 |
|  | 20 | 1500 | 10.86099 | 0.7496 | 1.3517 |
|  | 30 | 2000 | 22.52393 | 4.2265 | 4.9781 |

TABLE 5. THE MEAN FITNESS VALUES FOR THE GRIEWANK FUNCTION

| $N$ | $D$ | $T$ | FPSO [14] | DPSO [19] | SOPSO |
|---|---|---|---|---|---|
| 20 | 10 | 1000 | 0.091623 | 0.06506 | 0.06100 |
|  | 20 | 1500 | 0.027275 | 0.02215 | 0.02365 |
|  | 30 | 2000 | 0.02156 | 0.01793 | 0.02326 |
| 40 | 10 | 1000 | 0.075674 | 0.05673 | 0.05251 |
|  | 20 | 1500 | 0.031232 | 0.02150 | 0.01783 |
|  | 30 | 2000 | 0.012198 | 0.01356 | 0.01271 |
| 80 | 10 | 1000 | 0.068323 | 0.05266 | 0.04659 |
|  | 20 | 1500 | 0.025956 | 0.02029 | 0.01932 |
|  | 30 | 2000 | 0.014945 | 0.01190 | 0.01034 |

Table 3 to 5 lists the mean fitness values for three functions. It is easy to see that SOPSO have better results than FPSO [14] for almost all cases. By compare it with the results of DPSO [15], SOPSO also performs better for all the cases of Rosenbrock function, and better than most cases of Griewank function slightly, although it performs worse for the cases of Rastrigrin function.

### B. Device Optimization Example

The performance of SOPSO was also demonstrated on a double-implantation focused-ion-beam MOSFETs [13, 21], as shown in Fig. 5. A device simulator PISCES-2ET [22] is used to calculate the device characteristics. Here most parameters are fixed. The effective channel length ($L_{eff}$) is 0.25μm; the oxide thickness ($T_{ox}$) is 0.01μm. For source and drain, the junction depth ($X_j$) is 0.1μm, doping concentration ($N_{SD}$) is 7.0E20cm$^{-3}$. For both of the P$^+$ implant peaks in the channel, there have same implant energy as 10keV.

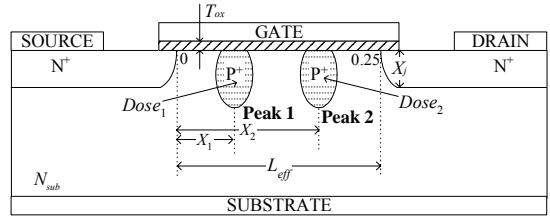

Figure 5. Schematic of 0.25μm double-implantation MOS device.

TABLE 6. DEVICE SEARCH SPACE AND PARAMETER PRECISIONS

| Name | Min | Max | σ | Unit |
|---|---|---|---|---|
| $X_1$ | 0.00 | 0.25 | 2.5E-4 | μm |
| $Dose_1$ | 1E10 | 1E13 | 1E10 | cm$^{-2}$ |
| $X_2$ | 0.00 | 0.25 | 2.5E-4 | μm |
| $Dose_2$ | 1E10 | 1E13 | 1E10 | cm$^{-2}$ |
| $N_{sub}$ | 1E15 | 1E18 | 1E15 | cm$^{-3}$ |

The design parameters include lateral implantation position that start from source side of channel for peak 1

($X_1$) and 2 ($X_2$), implantation dose in peak 1 ($Dose_1$) and 2 ($Dose_2$), and substrate doping concentration ($N_{sub}$). Table 6 lists the lower and upper boundary value and the precision $\sigma$ for each parameter, which includes almost all the possible implantation states in the device channel.

Table 7 lists the objectives, which includes drive current ($I_{on}$) and output conductance ($G_{out}$) at $V_{ds}$=1.5V, $V_{gs}$=1.5V; and off current ($I_{off}$) at $V_{ds}$=1.5V and $V_{gs}$=0V.

TABLE 7. OBJECTIVES: DESIRED DEVICE PERFORMANCE

| Name | Objective | Unit |
|---|---|---|
| $I_{on}$ | maximize | A/μm |
| $I_{off}$ | ≤ 1E-14 | A/μm |
| $G_{out}$ | ≤ 8E-6 | 1/Ω |

Several algorithm settings are tested. GENOCOP is a real-value genetic algorithm (GA) with multiply genetic operators [9], which had applied for optimizing device successfully [8, 21]. Here its population size is set as $N_{pop}$=50. For each generation, it has $N_c$=10~12 children individuals, the selection pressure $q$=0.01. For standard PSO, $w$ are fixed as 1 [7], 0.4, and a linearly decreasing $w$ which from 0.9 to 0.4 [14], respectively. For SOPSO, $w$ is fixed as 0.4. The number of particles $N$=10 for all the PSO versions. We had 20 trial runs for every instance.

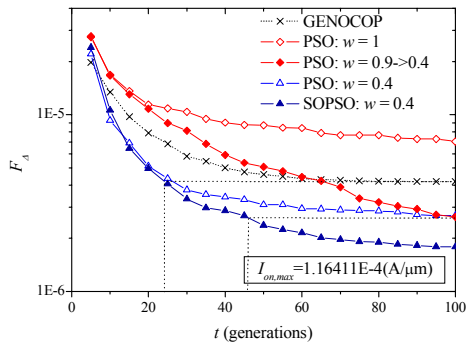

Figure 6. Relative mean fitness $F_\Delta$ of device optimization results.

Fig. 6 shows the relative mean performance $F_\Delta$=|$F_{opt}$-$F_B$| during 100 generations for different algorithms. Where $F_B$ is the mean fitness value of the best particle found in current generation that found by algorithm, and $F_{opt}$=1.16411E-4A/μm is the optimum value for $I_{on}$ when satisfying the constraints on $I_{off}$ and $G_{out}$. For all the PSO versions, the total evaluation times is $T_e$=1000, which is a little less than that of GENOCOP (about 1050~1250). It shows that the original PSO version with $w$=1 [7], performed worst in all cases since it is worked in chaotic state. However, all the other PSO versions perform better than GENOCOP. The PSO version with $w$=0.4 converged fastly and stagnated at the last stage of evolution, since it is worked in dissipative state. The PSO version with a linearly decreasing $w$ which from 0.9 to 0.4 performed better than the PSO version with $w$=0.4 at last which converged slowly at the early stage while fastly at the last stage,. Moreover, the SOPSO performed as the best in all cases, which evolving sustainable when the evolution of PSO with same $w$ is going to be stagnated. In addition, SOPSO costed only 24 generations to achieve the performance of GENOCOP, and only 46 generations to achieve the performance of PSO with $w$ =0.4.

VII. CONCLUSION

In this paper, a self-organizing PSO was presented by simulating the self-organization. The particle swarm is worked in dissipative condition by employing a small $w$, based on experimental analysis for a simplified model, which with fast convergence. Then the negative entropy is introduced, which is realized by recognizing and replacing inactive particles according to the existing information of the problem, i.e. the process deviations of device parameters, so as to driving the irreversible evolution process with better fitness, while removes the additional control parameters.

The testing of three benchmark functions indicates the SOPSO has good performance. Then the testing results on a FIBMOS device illustrate that the SOPSO has fast optimization capability from the early searching stage, which is crucially for device optimization since each evaluation by device simulation is time-consumption.